\documentclass{article}

\usepackage{arxiv}

\usepackage[utf8]{inputenc} 
\usepackage[T1]{fontenc}    
\usepackage{hyperref}       
\usepackage{url}            
\usepackage{booktabs}       
\usepackage{amsfonts}       
\usepackage{nicefrac}       
\usepackage{microtype}      
\usepackage{lipsum}		
\usepackage{graphicx}
\usepackage{natbib}

\usepackage{doi}

\title{Exploiting Meta-Cognitive Features for a Machine-Learning-Based One-Shot Group-Decision Aggregation}

\date{} 					

\author{
  Hilla Shinitzky, Yuval Shahar, Dan Avraham, Yizhak Vaisman, Yakir Tsizer and Yaniv Leedon \\
  The Department of Software and Information Systems Engineering  \\
  Ben-Gurion University of the Negev \\
  Beer-Sheva, Israel\\
  Corresponding author: \texttt{hillash@post.bgu.ac.il}\\
}

\renewcommand{\headeright}{}
\renewcommand{\undertitle}{}

\hypersetup{
pdftitle={Exploiting Meta-Cognitive Features for a Machine-Learning-Based One-Shot Group-Decision Aggregation},
}

\begin{document}
\maketitle

\begin{abstract}

The outcome of a collective decision-making process, such as crowdsourcing, often relies on the procedure through which the perspectives of its individual members are aggregated. Popular aggregation methods, such as the majority rule, often fail to produce the optimal result, especially in high-complexity tasks. Methods that rely on meta-cognitive information, such as confidence-based methods and the Surprisingly Popular Option, had shown an improvement in various tasks. However, there is still a significant number of cases with no optimal solution. Our aim is to exploit meta-cognitive information and to learn from it, for the purpose of enhancing the ability of the group to produce a correct answer. Specifically, we propose two different feature-representation approaches: (1) \textit{Response-Centered feature Representation} (RCR), which focuses on the characteristics of the individual response instances, and (2) \textit{Answer-Centered feature Representation} (ACR), which focuses on the characteristics of each of the potential answers. Using these two feature-representation approaches, we train \textit{Machine-Learning} (ML) models, for the purpose of predicting the correctness of a response and of an answer. The trained models are used as the basis of an ML-based aggregation methodology that, contrary to other ML-based techniques, has the advantage of being a "one-shot" technique, independent from the crowd-specific composition and personal record, and adaptive to various types of situations. To evaluate our methodology, we collected 2490 responses for different tasks, which we used for feature engineering and for the training of ML models. We tested our feature-representation approaches through the performance of our proposed ML-based aggregation methods. The results show an increase of 20\% to 35\% in the success rate, compared to the use of standard rule-based aggregation methods.

\end{abstract}

\section{Introduction}

Combining choices made by multiple individuals has been shown to improve the quality of the final collective outcome, relative to the average performance of an individual (i.e., the "wisdom of the crowd" phenomenon, popularized by \citet{surowiecki2005wisdom}), in various tasks. Thus, the optimal final decision can be accurately aggregated from a collection of decisions through a simple \emph{majority rule} (i.e., choosing the option that the majority agrees on), in a non-negligible portion of the time.
As described in the Condorcet Jury Theorem \citep{de1785essai}, for binary-choice decisions, made by a collective, as long as the probability of an individual to choose the correct answer is higher than 0.5, the probability to reach a correct collective decision grows with the number of individuals.

However, in some cases, such as in high complexity tasks in which the individual's probability of finding the optimal solution is low, the majority decision can be often incorrect 
\citep{bahrami2010optimally,chen2004eliminating,koriat2018prototypical,lorenz2011social,simmons2010intuitive}.  This phenomenon holds true even if a second chance to select the correct answer is provided, by enabling each group member to consider correct and incorrect answers suggested by the other group members, when it is difficult for an individual who had chosen the wrong answer to be persuaded when looking at a correct answer, due to the difficulty of verifying its validity
\citep{amir2013verification,amir2018more}.
Thus, there are multiple scenarios in which a correct collective answer is especially challenging to achieve. 

Though methods for aggregating the final decision of a collective, that rely on meta-cognitive information (e.g., confidence-based methods), have been proven to be successful in some cases, there are still a significant amount of failures \cite{bahrami2010optimally,koriat2016views,lee2012inferring,chen2004eliminating,koriat2018prototypical,lorenz2011social,simmons2010intuitive,levine2015social}.

We offer an approach to exploit and learn from the meta-cognitive information, as we attempt to answer the following question: Are engineered features based on meta-cognitive data able to explain and predict the performance of a worker and the correctness of an answer?
Specifically, we propose, implement and evaluate two different feature-representation approaches:
(1) Response-Centered feature Representation (RCR), which focuses on the characteristics of the individual response instances (e.g., the specific response instance's confidence),
and (2) Answer-Centered feature Representation (ACR), which focuses on the characteristics of each of the potential answers (e.g., mean actual or predicted support by the group members).

At the core of our approach are two classification models and their corresponding set of engineered features. The first predicts the worker's performance on a specific task, meaning the correctness of a specific response. The second predicts the correctness of an answer to a specific task, using aggregated data from its supported responses.
By applying explain-ability techniques, we observe the impact of each feature on models output and infer how useful each meta-cognitive information is for aggregation procedures. Thus, we can achieve a deeper understanding of what characterizes a correct answer and what makes a correctly solving worker. 

The learned models are used for new ML-based aggregation methods that, given a set of responses, infer the decision that is most likely to be correct, by (1) applying a \textit{feature engineering} process, (2) using several different machine-learning models to train (induce) one or more classifiers, and (3) applying the trained classifiers to new, hitherto unseen, group-decision data, to determine the correct decision. 
Contrary to other Machine-Learning based aggregation methods, our proposed methods has the advantage of being a "one-shot" technique, independent from the crowd specific composition, prior knowledge, history or personal record, and adaptive to various types of situations. 

To evaluate our approach, we collected data by conducting online studies — each study included one problem-solving task, followed by questions eliciting meta-cognitive and social meta-cognitive information. 
Experimental results indicate our methods' ability to identify the correct answer, even in cases where all other techniques fail. Our methods are likely to be useful for any group facing a complex strategic problem, or for implementing a successful answer-aggregation for collective tasks. 

Although our current approach is applied to several specific decision-making domains, we presume that there is a great potential for a significant contribution to crowdsourcing, group decision-making, and collective-intelligence research by examining this idea from a broader perspective. We conjecture that, by collecting and learning over empirical data that include social meta-cognitive aspects, we can obtain valuable knowledge that could be vital information for the intelligent solution aggregation process.

\section{Related Work}
\label{sec:headings}

Crowdsourcing is a good demonstration for a non-interacting group of individuals who work in parallel on large tasks. Examples include weather forecasting \citep{muller2015crowdsourcing}, labeling data for supervised learning, and medical diagnosis \citep{raykar2010learning,kurvers2016boosting}.

However, there are still types of tasks in which the collective does not surpass individuals' performance, and in some cases, can even become less effective as the size of the group grows, which could be due to a variety of reasons, such as the problems' computational characteristics, and the expected success of an individual \citep{bahrami2010optimally,chen2004eliminating,koriat2018prototypical,lorenz2011social,simmons2010intuitive,amir2013verification,amir2018more}. 

Focusing on nominal groups or crowds, a core element in the attempts to improve the quality of the outcome is the mechanism through which the set of individuals' choices are assembled into one collective decision. 
The most common aggregation method, which leans on the wisdom of crowd principle \citep{surowiecki2005wisdom}, is the majority rule (MR), which takes the decision that the majority of individuals agrees on. This method can be implemented easily and often aggregates correctly a set of decisions \citep{kerr2004group,sorkin1998group}. 

In tasks that require some expertise or specific knowledge, it could be better to be more selective with the individuals on whom the final decision relies on. 
For instance, some studies examined the possibility of identifying experts and give those more weight when aggregating the final group decision \citep{budescu2014identifying,lee2012inferring,yue2014weighted}. Applying the same notion but in more extreme fashion, some studies suggested aggregating the final decision by taking only the perspective of a selected few in the group \citep{goldstein2014wisdom,mannes2014wisdom}. However, such methods require some information about the individuals, such as historical records on past performance.

Methods that rely on some individuals more than others were also applied through subjective-confidence, i.e., the reported confidence of the individual about their decision. These methods are implemented through weighting (i.e., weight the individual's decision in the collective decision according to their reported confidence) and maximization (i.e., choose the decision with the highest average confidence reported by its supporters). 
Although it has been shown that the subjective-confidence-based methods can work in some cases \citep{aydin2014crowdsourcing,koriat2008subjective,koriat2012two}, these approaches are not always dependable and suited for specific domains \citep{bahrami2010optimally,hertwig2012tapping,koriat2016views,lee2012inferring}. 
This implies that the subjective confidence of an individual is an important feature to consider, but better to do so while taking into account the context of the decision, especially for constructing a generalized aggregation approach. The use of Machine-Learning techniques can provide the flexibility needed here \citep{laan2017rescuing}.

Another example for the use of meta-cognitive data, in addition to the use of the subjective confidence, is the surprisingly popular option (SP) method \citep{prelec2017solution}. The SP method chooses the answer that has had a surprisingly large support, relative to the mean predicted support of that answer, as estimated by the participants and not necessarily the majority’s opinion. The method fared well in the researchers’ experiments, though mostly in common-knowledge binary questions.

Strategic multiple choice problems often require more sophisticated and suitable approaches, and there are methods specifically for aggregating the crowd’s answers to multiple choice questions. One example is the aggregation method developed by \citet{yi2012wisdom}, which creates a solution composed by popular solution pieces from a combination of individuals' solutions.

These techniques for aggregating the final decision of a collective, can often be summed up as using a simple, and sometimes effective, deterministic aggregation rule. 
The use of sophisticated model-based methods \citep{bachrach2012grade,bachrach2012crowd,zheng2017truth}, statistical and Machine-Learning methods \citep{gaunt2016training,laan2017rescuing,weld2015artificial} has been attempted in the context of collective decision-aggregation and crowdsourcing systems. Our approach differs from these methods, first by its unique meta-cognitive features engineering combined with data science techniques. Moreover, while previous Machine-Learning based aggregation methods use matrices of responses and performance history of each individual, our methodology has the advantage of being a “one-shot” technique. Meaning, it is independent of the crowd-specific composition, prior knowledge, history or personal record of decision making, and is adaptive to various types of situations (contexts) in automated fashion.

Given the techniques and insights provided from previous studies, it has become clear that there is a variety of important meta-cognitive features that need to be considered, in addition to the simple, standard support for each answer to the original decision problem, such as subjective confidence and predicted distribution of answers' support; and that it might also be important to consider the context of the decision (e.g., the complexity of the problem, the uncertainty expressed by the group members who try to solve the decision problem, etc.). 

Thus, to provide the necessary flexibility, we propose to construct a generalized aggregation approach, suited for a broad range of cases, based on Machine-Learning techniques, by learning (given multiple collective decision-making instances) the relationship between patterns composed of simple collective's features, meta-cognitive features, and new features derived from them, and the optimal answers to the associated problems. Our approach exploits the flexibility of Machine-Learning techniques for creating a robust aggregation method that attempts to handle multiple types of challenging decision-aggregation cases. 

\section{Definitions and Methods}

We focus on a domain of problems, where there are number of options on the table to choose from (i.e., multiple-choice), and only one is considered to be the optimal choice, or correct solution.
Formally: for a given problem $P$ with a number of possible answers $m$, let $A_P$ denote the set of answers, $\forall{i\in(1,\dots,m)}|a_i\in A_P$, i.e., $a_i$ is a possible answer to the problem $P$, and there is (exactly) one optimal answer, denoted by $a^*$. 
The collection of opinions regarding $P$, is given by the set of responses $R_P$, where each $r\in R_P$ is a response of one individual. 

In the basic form of a response $r$, each response contains a vote for one answer in $A_P$.
We define an aggregation method as a function, which for a given $R_P$, returns an answer $a$ to be the collective's aggregated answer to the problem $P$, based on the method's rule or criteria. 
The primary goal of an aggregation method is to maximize its capabilities to identify the optimal answer, i.e., maximize the chance that the returned answer will be $a^*$. An optimal aggregation method would return $a^*$ in $100\%$ of the cases.

Thus, our input data, is a set of responses $R_P$, where each response $r\in R_P$ contains:
\begin{itemize}
\item $v$ - a vote for an answer in $A_P$; 
\item $\left \{ ps[a_1],...,ps[a_m] \right \}$ - Predicted support for each $a_i \in A_P$;
\item $c$ - reported confidence.
\end{itemize}  

Before describing our proposed methods, we first lay down basic definitions.

\begin{description}
	\item[Supporters and non-supporters of answer $a$] We refer to the sub-set $R_{P(a)}\subseteq R_P$ as the \emph{supporters} of answer $a$, i.e., the responses of those who voted for $a$ (formally, $R_{P(a)}=\{r_j \in R_P | r_j.v=a\}$). In a similar way, we refer to the $\overline{R_{P(a)}}$ as the \emph{non-supporters} of $a$, i.e., the responses of those who voted for any answer in $A_P$ except $a$ (formally, $\overline{R_{P(a)}}=R_P\setminus R_{P(a)}$).
	\item[In-group and Out-group of response $r$] The sub-set $R_{P(r.v)}$ is referred to as the \emph{in-group} of $r$ and the sub-set $\overline{R_{P(r.v)}}$ is referred to as the \emph{out-group} of $r$. 
	\item[Support Rate of answer $a$] $S(a)$ denotes the (actual) support rate of $a$ (i.e., the percentage of votes) in a given set of responses. This can be calculated by $S(a) = \frac{|R_{P(a)}|}{|R_P|}$.
    \item[Support Distribution] $S = \{ S(a_i) , \forall{i\in(1,\dots,m)}|a_i\in A_P\}$ denotes the (actual) support rate distribution for the problem $P$, given a set of responses $R_P$. 
    \item[Solver] A response $r$ is referred to as a \emph{solver}'s response iff $r.v=a^*$ (i.e., the voted answer is the correct one).
    \item[Average Predicted Support] A function $AvgPS(R_P,a)$ returns the average predicted support of a given answer $a$ in a given set of responses $R_P$. 
\end{description}

Next, we define two feature-representation approaches.

\subsection{Response-Centered feature Representation}

The \textit{Response-Centered feature Representation} (RCR) feature-engineering approach focuses on the characteristics of the individual response instances. For example, the individual's confidence in their answer, or its deviation from the mean confidence of the responses of the group.

Table \ref{tab:rcr-features} details the features of an individual response instance, based on the original response $r$, originated from a given set of responses $R_P$. 
The composed features include: (1) Basic features: raw data from the associated responses, unprocessed (e.g., \textit{Conf}, \textit{PSv}); (2) Features representing the information, used by the standard rule-based aggregation method (e.g., \textit{ChosenMajAns} for MR, \textit{dPSv} for SP); (3) Features comparing the predicted vs actual answers' support distribution (e.g., \textit{ChiSq}); (4) Features comparing the response's information to the overall average in $R_P$ (e.g., \textit{PSvD}, \textit{ConfD}, \textit{PChiSq}); (5) For every feature extracted by comparing the response to all other responses, we extracted two more features, which express the same information but with respect to the response's \textit{in-group} (e.g., \textit{PSvD${_{IG}}$}, \textit{ConfD${_{IG}}$}, \textit{PChiSq${_{IG}}$}) and its \textit{out-group} (e.g., \textit{PSvD${_{OG}}$}, \textit{ConfD${_{OG}}$}, \textit{PChiSq${_{OG}}$}).

Note that we do not assume any record of past performance of the respondent, or any identifiable features of the actual decision-maker; Also, there are no features referring to the actual task, nor to the answers' contents.

\subsubsection{Training, Classification, and Aggregation based on the Response-Centered feature Representation}

The first Machine Learning methodology we use, aggregates the results of the group's votes by exploiting the RCR feature-engineering approach; we refer to it as the \textit{RCR-Agg method}. This method classifies each of the instances of \textit{responses} to the original problem as being True (correct) or False (incorrect), and then performs an aggregation operation on the classified responses to determine the final answer to the original problem that the group was faced with.

We are assuming that the input to the RCR-Agg method is a set of sets of responses, each produced by a group attempting to solve some given problem (possibly even the same problem). Each response is labeled as being [eventually] True (correct) or False (incorrect).

The training, classification, and aggregation process, as performed by the RCR-Agg method, includes the following steps: 

\begin{enumerate}
    \item Generate, for each response instance from the $n$ sets of responses ${R_{P_1},..,R_{P_n}}$, a set of RCR features.
    \item Train a classifier to classify a given response into True (correct) or False (incorrect) using the RCR features.
    \item Given a new set of responses $R_{P_{new}}$, obtain classification predictions for each response instance within $R_{P_{new}}$, using the RCR-based classifier.
    \item Finally, return, as the suggested answer for the $P_{new}$ problem, an answer based on an aggregation of the all of the classifications of the individual response instances within $R_{P_{new}}$. We examined five different aggregation strategies, given the classification labels and the predictions' probabilities (i.e., the probability of the classified instance to be True, according to the classifier):
    \begin{itemize}
        \item [\textit{maj}]: Return the answer whose number of "True" classified supporters is the largest (i.e., the majority vote, within the group of  response instances classified as "True"). 
        \item [\textit{prop}]: Return the answer with the highest proportion of "True" versus "False" classified supporters.
        \item [\textit{wm}]: Return the answer with the highest \textit{sum} of supporters' predictions' probabilities (i.e., a weighed majority vote). 
        \item [\textit{avgp}]: Return the answer with the highest \textit{average} of supporters' predictions' probabilities.
        \item [\textit{maxp}]: Return the answer with the highest \textit{maximum value} of supporters' predictions' probabilities.
    \end{itemize}
    Note that for the methods \textit{maj} and \textit{prop}, we use one of the other three methods (\textit{wm}, \textit{avgp} and \textit{maxp}) as a tie breaker, if needed. Thus, considering the different tie breaker options, there are 9 classification aggregation strategies, in total. 
\end{enumerate}

\begin{table}
    \caption{Response-Centered feature Representation (RCR): Features Table}
    \centering
    \begin{tabular}{lll}
	    \toprule
		\cmidrule(r){1-2}
		Feature Name     & Value     & Description \\
		\midrule
        \textit{ChosenMajAns}                                                                                                       & $\{1,0\}$         & \begin{tabular}{@{}l@{}}1:if $r.v$ is the majority answer \\ 0:else\end{tabular}                     \\ 
        \textit{Conf}                                                                                                     & $r.c$          & Reported  confidence     \\ 
        \textit{ConfD}                                                                                                     & $r.c-\frac{\sum_{r_j\in R_P}^{}(r_j\cdot c)}{|R_P|}$          & \begin{tabular}{@{}l@{}}Confidence distance from all \\ responses’ average\end{tabular}   \\ 
        \textit{ConfD$_{IG}$}                                                                                                     & $r.c-\frac{\sum_{r_j\in R_{P(r.v)}}^{}(r_j\cdot c)}{|R_{P(r.v)}|}$          & \begin{tabular}{@{}l@{}}Confidence distance from \\ in-groups’ average\end{tabular}     \\ 
        \textit{ConfD$_{OG}$}                                                                                                    & $r.c-\frac{\sum_{r_j\in \overline{R_{P(r.v)}}}^{}(r_j\cdot c)}{|\overline{R_{P(r.v)}}|}$          &  \begin{tabular}{@{}l@{}}Confidence distance from \\ out-groups’ average\end{tabular}  \\ 
        \textit{PSv}                                                                                                     & $r.ps[r.v]$          &    \begin{tabular}{@{}l@{}}Predicted support of v \\ (chosen answer)\end{tabular}   \\    
        \textit{dPSv}                                                                                                     & $S(r.v)-r.ps[r.v]$          & Prediction error of v’s support  \\    
        \textit{PSvD}                                                                                                     & $r.ps[r.v]-AvgPS(R_P,r.v)$          &     \begin{tabular}{@{}l@{}}The distance of predicted support \\ of v from the average prediction \\ of all responses\end{tabular} \\   
        \textit{PSvD$_{IG}$}                                                                                                   & $r.ps[r.v]-AvgPS(R_{P(r.v)},r.v)$          &   \begin{tabular}{@{}l@{}}The distance of predicted support \\ of v from the average prediction \\ of the in-group\end{tabular} \\    
        \textit{PSvD$_{OG}$}                                                                                                    & $r.ps[r.v]-AvgPS(\overline{R_{P(r.v)}},r.v)$          &    \begin{tabular}{@{}l@{}}The distance of predicted support \\ of v from the average prediction \\ of the out-group\end{tabular} \\  
        \textit{ChiSq}                                                                                & $\frac{1}{2} \cdot \sum_{a_i \in A_P}^{} \left ( \frac{(r.ps[a_i]-S(a_i))^2}{r.ps[a_i]+S(a_i)} \right )$          &    \begin{tabular}{@{}l@{}}Chi-square score: statistic of \\ chi-square distributions comparison test \\ (predicted vs actual answer distribution) \end{tabular}  \\     
        \textit{PChiSq}                                                                                                    & $\frac{1}{2} \cdot \sum_{a_i \in A_P}^{} \left ( \frac{(r.ps[a_i]-AvgPS(R_P,a_i))^2}{r.ps[a_i]+AvgPS(R_P,a_i)} \right )$          &    \begin{tabular}{@{}l@{}}Chi-square score for predicted vs \\ distribution of the average prediction \\  of all responses   \end{tabular}  \\            
        \textit{PChiSq${_{IG}}$}   & $\frac{1}{2} \cdot \sum_{a_i \in A_P}^{} \left ( \frac{(r.ps[a_i]-AvgPS(R_{P(r.v)},a_i))^2}{r.ps[a_i]+AvgPS(R_{P(r.v)},a_i)} \right ) $         &  \begin{tabular}{@{}l@{}}Chi-square score for predicted vs \\ distribution of the average prediction \\  of the in-group   \end{tabular}  \\            
        \textit{PChiSq${_{OG}}$}
            & $\frac{1}{2} \cdot \sum_{a_i \in A_P}^{} \left ( \frac{(r.ps[a_i]-AvgPS(\overline{R_{P(r.v)}},a_i))^2}{r.ps[a_i]+AvgPS(\overline{R_{P(r.v)}},a_i)} \right ) $         &  \begin{tabular}{@{}l@{}}Chi-square score for predicted vs \\ distribution of the average prediction \\  of the out-group   \end{tabular} \\
	\bottomrule
\end{tabular}
   \label{tab:rcr-features}
\end{table}

\subsection{Answer-Centered feature Representation}

The \textit{Answer-Centered feature Representation}(ACR) feature-engineering approach focuses on the characteristics of each of the potential answers, i.e., the various \textit{Answers} offered to the respondents. The base data-item instance here is an \textit{answer} to a problem-solving task and its features, extracted from the responses to the problem-solving task that the answer addresses. Note that the features of the answers' data-item instances, such as the mean actual support of that answer by the group, or the mean \textit{predicted} support to that answer (a meta-cognitive feature), do not include the problem-solving task that each answer refers to. 

Table \ref{tab:acr-features} details the features of an answer's instance, extracted from a given set of responses $R_P$. The feature-engineering process here, is similar, in nature, to the one described for RCR. 
For example, here we also apply one function on different sub-sets of responses, to extract different features; similar to the \textit{in-group} and \textit{out-group} sub-sets division applied for responses, the parallel division applied here is of \textit{supporters} and \textit{non-supporters} of the answer (since the point of reference here is answers rather than responses) - e.g., \textit{AvgPS${_{IG}}$}, \textit{AvgPS${_{OG}}$}, \textit{dPS${_{IG}}$}, \textit{dPS${_{OG}}$}. 
The composed features also include: 
(1) Features representing the information, used by the standard rule-based aggregation method (e.g., \textit{IsMajority} for MR, \textit{dPS} for SP); 
(2) Features based on the actual answers' support distribution (e.g., \textit{Support}); 
(3) Feature extracted from average values of responses' features, associated with the answers' supporters (e.g., \textit{AvgConf}, \textit{AvgChisq}).

\subsubsection{Training, Classification, and Aggregation based on the Answer-Centered feature Representation}

The second Machine Learning method we use, aggregates the results of the group's votes by exploiting the ACR feature-engineering approach; we refer to it as the \textit{ACR-Agg method}. The ACR-Agg method classifies each of the set of possible \textit{answers} to the original problem as being either True (correct) or False (incorrect). It then determines which answer is the correct one, based on the classification results of all potential answers, using an aggregation procedure.

We are assuming that the input to the ACR-Agg method is again a set of sets of responses ${R_{P_1},..,R_{P_n}}$, each set being produced by a group attempting to solve some given problem (possibly even the same problem). Note that each response contains a vote for one answer out of the answer set $A_p$. Recall that each answer set might have a different number of answers, $m$.

However, this time, each \textit{answer} is (eventually) labeled as being  True (correct) or False (incorrect).

The training, classification, and aggregation process, as performed by the ACR-Agg method, includes the following steps: 

\begin{enumerate}
    \item Generate, for each single answer out of the set of sets of answers ${A_{P_1},..,A_{P_n}}$, each set of answers being of a size $m_i$, ${i\in(1,\dots,n)}$, a set of ACR features, using the sets of responses ${R_{P_1},..,R_{P_n}}$.
    \item Train a classifier to classify a given answer into True (correct) or False (incorrect) using its ACR features.
    \item Given a new set of responses $R_{P_{new}}$, for which the answer set is $A_{P_{new}}$, obtain classification predictions for each answer instance within $A_{P_{new}}$, using the ACR-based classifier.
    \item Finally, return, as the suggested answer for the $P_{new}$ problem, an answer based on an aggregation of the all of the classifications of the individual answer instances within $A_{P_{new}}$.
    To aggregate \textit{answer} classifications, we used the following aggregation strategy:
    There are three mutually exclusive cases to be considered, according to the results of the classification process: 
    \begin{itemize}
        \item If only one answer is classified as "True", this answer is selected;
        \item If multiple answers are classified as "True", the answer with the highest prediction probability is selected; 
        \item If all answers are classified as "False", the answer with the lowest predicted probability of being False is selected.
    \end{itemize}  
\end{enumerate}

\begin{table}
    \caption{Answer-Centered feature Representation (ACR): Features Table}
    \centering
    \begin{tabular}{lll}
	    \toprule
		\cmidrule(r){1-2}
		Feature Name     & Value     & Description \\
		\midrule
        \textit{IsMajority}                                                                                                       & $\{1,0\}$         & \begin{tabular}{@{}l@{}}1: if a is the majority answer
         \\ 0 : else \end{tabular}                      \\ 
        \textit{Support}                                                                                                     & $S(a)=\frac{|R_{P(a)}|}{|R_{P}|}$          & Answers’ support
            \\ 
        \textit{AvgPS}                                                                                                     & $AvgPS(R_P,a)=\frac{\sum_{r_j\in R_P}^{}(r_j\cdot ps[a])}{|R_P|}$          & \begin{tabular}{@{}l@{}}Average predicted support  \\ by all responses \end{tabular}      \\ 
        \textit{AvgPS$_{IG}$}   & $AvgPS(R_{P(a)},a)=\frac{\sum_{r_j\in R_{P(a)}}^{}(r_j\cdot ps[a])}{|R_{P(a)}|}$          & \begin{tabular}{@{}l@{}}Average predicted support  \\ by supporters \end{tabular}  \\ 
        \textit{AvgPS$_{OG}$}     & $AvgPS(\overline{R_{P(a)}},a)=\frac{\sum_{r_j\in \overline{R_{P(a)}}}^{}(r_j\cdot ps[a])}{|\overline{R_{P(a)}}|}$          & \begin{tabular}{@{}l@{}}Average predicted support  \\ by non-supporters \end{tabular}   \\ 
        \textit{dPS}                                                                                                     & $S(a)-AvgPS(R_P,a)$          & \begin{tabular}{@{}l@{}}Predicted support error by   \\ all responses  \end{tabular}  \\    
        \textit{dPS$_{IG}$}                                                                                                    & $S(a)-AvgPS(R_{P(a)},a)$          & \begin{tabular}{@{}l@{}}Predicted support error by   \\ supporters \end{tabular}  \\    
        \textit{dPS$_{OG}$}                                                                                                     & $S(a)-AvgPS(\overline{R_{P(a)}},a)$          & \begin{tabular}{@{}l@{}}Predicted support error by   \\ non-supporters \end{tabular}  \\    
        \textit{AvgChiSq}                                                                                                  & $\frac{\sum_{r_j\in R_{P(a)}}^{}(ChiSq(r_j))}{|R_{P(a)}|}$          & \begin{tabular}{@{}l@{}}Average supporters’ chi    \\ square score \end{tabular}  \\    
        \textit{AvgConf}                                                                                                   & $\frac{\sum_{r_j\in R_{P(a)}}^{}(r_j\cdot c)}{|R_{P(a)}|}$          & Average supporters’ confidence 
          \\  
	\bottomrule
\end{tabular}
   \label{tab:acr-features}
\end{table}

\section{Evaluation Procedure}

For data collection, we have conducted several online experiments via Amazon Mechanical Turk (Amazon Co.) crowd-sourcing platform. 
We chose to experiment on 22 different problem-solving tasks; 14 logic riddles and misleading math problems (4 were taken from \citet{ackerman2014diminishing}), and 8 game formed tasks of P-space complete and NP-complete decision and optimization problems (e.g., "Rush-Hour" puzzle \citep{flake2002rush} and the Knapsack problem \citep{mathews1896partition}). 
Each online form included demographic questionnaire (e.g., age, education, gender, etc.) and a problem-solving task (in a form of a multiple-choice question), followed by (1) confidence report: \textit{“How confident are you in your answer?”}; (2) prediction regarding other respondents' answers to the associated problem \textit{“Out of the people trying to solve this problem, what percentage do you estimate, will select each of the available options?"}. 
We collected responses for each problem-solving task, and used these data to evaluated our approach. In this paper, we present the results of 2490 responses (for all 22 problem-solving tasks combined). 

The online experiments produced a set of responses $R_P$ for each problem $P$. The next phase to generate data sets, used for training and testing the responses-based and answers-based classification models. 
Prior to the training and evaluation phase, we have conducted a process of bagging and sampling, on which we have sampled a fixed-size sub-set of responses. After experimenting with different sizes (15, 20, 25, 30 and 35) of responses' sub-sets, we found 30 to be an appropriate virtual-group size to sample. Future work can focus on developing a generalized approach to determine the optimal sub-set sizes under different conditions.

We evaluated our two feature representation approaches (RCR and ACR) through training ML models using the described data sets, and applying feature-evaluation techniques, and specifically through the computation of the SHapely Additive exPlanation (SHAP) values \citep{lundberg2017unified}.

The evaluation process of our ML-based aggregation methods, RCR-Agg and ACR-Agg, was conducted in a form of "leave one group out" procedure, as follows: For each sub-set $R_P'$, we trained the model over the data generated from all of the remaining sub-sets of responses, and applied it to the left-out sub-set; we averaged the resulting performance over all such subsets. 

The results presented in this paper are of the best performing classification technique, a weighted average probabilities voting ensemble technique (i.e., soft voting) \citep{breiman1996bagging,wolpert1992stacked,freund1995boosting,dietterich2000ensemble} which is composed of five classifiers: Linear Discriminant Analysis, Random Forest, XGboost, Logistic Regression, and K-Nearest Neighbor \citep{chen2016xgboost,bishop2006pattern,breiman2001random,menard2002applied}.
The results of the RCR-Agg method presented in this paper were obtained by applying the classification aggregation strategy \textit{prop} and the \textit{avgp} measure as the tie breaker; this was the best performing strategy among the nine different options that were described in the previous section. The complete results of all of the classification-aggregation strategies are shown in Table \ref{tab:class-agg-res}. 

\begin{table}[h]
    \caption{Classification Aggregation Strategies: Results Table}
    \centering
\begin{tabular}{llll}
	    \toprule
\textbf{Classification Aggregation Strategy} & \textbf{Tie Breaker} & \textbf{Success Rate} & \textbf{Percentage of Tie Breakers} \\
        \midrule
maxp                                         & -                    & 58\%                   & 0\%                         \\
avgp                                         & -                    & 61\%                   & 0\%                         \\
wm                                           & -                    & 47\%                   & 0\%                         \\
maj                                          & avgp                 & 62\%                   & 16\%                        \\
maj                                          & maxp                 & 60\%                   & 17\%                        \\
maj                                          & wm                   & 57\%                   & 16\%                        \\
prop                                         & avgp                 & 64\%                   & 18\%                        \\
prop                                         & maxp                 & 60\%                   & 16\%                        \\
prop                                         & wm                   & 57\%                   & 17\%                       \\
\bottomrule
\end{tabular}
\label{tab:class-agg-res}
\end{table}

We compared the results of RCR-Agg and ACR-Agg to the performance scores of other popular and commonly used methods: 

\begin{enumerate}
    \item \textbf{Majority-Rule (MR):} Returns the answer chosen by the majority of respondents, i.e., with the highest support. Formally, returns $a$ that maximizes the value: $S(a)$
    \item \textbf{Weighted-Confidence (WC):} Each vote is weighted by the respondent's confidence.  Returns the answer with the highest confidence-weighted support. Formally, returns $a$ that maximizes the value: $S(a)*\frac{\sum_{r_j\in R_{P(a)}}^{}(r_j\cdot c)}{|R_{P(a)}|}$ 
    \item \textbf{Highest Average Confidence (HAC):} Returns the answer with the highest average confidence reported by supporters. Formally, returns $a$ that maximizes the value: $\frac{\sum_{r_j\in R_{P(a)}}^{}(r_j\cdot c)}{|R_{P(a)}|}$ 
    \item \textbf{Surprisingly Popular (SP):} Returns the answer that was more popular than predicted. Formally, returns $a$ that maximizes the value: $S(a)-AvgPS(R_P,a)$
\end{enumerate}

\section{Results and Discussion}

As mentioned in the previous section, we used a combined total of 2490 responses (collected though online experiments) associated with 22 problems, from which we randomly sampled 77 sub-sets of $\sim30$ responses.
The average percentage of solvers was 28\%, indicating a high level of complexity in the problem-solving tasks.

\begin{figure}[h]
	\centering
  \includegraphics[width=0.7\columnwidth]{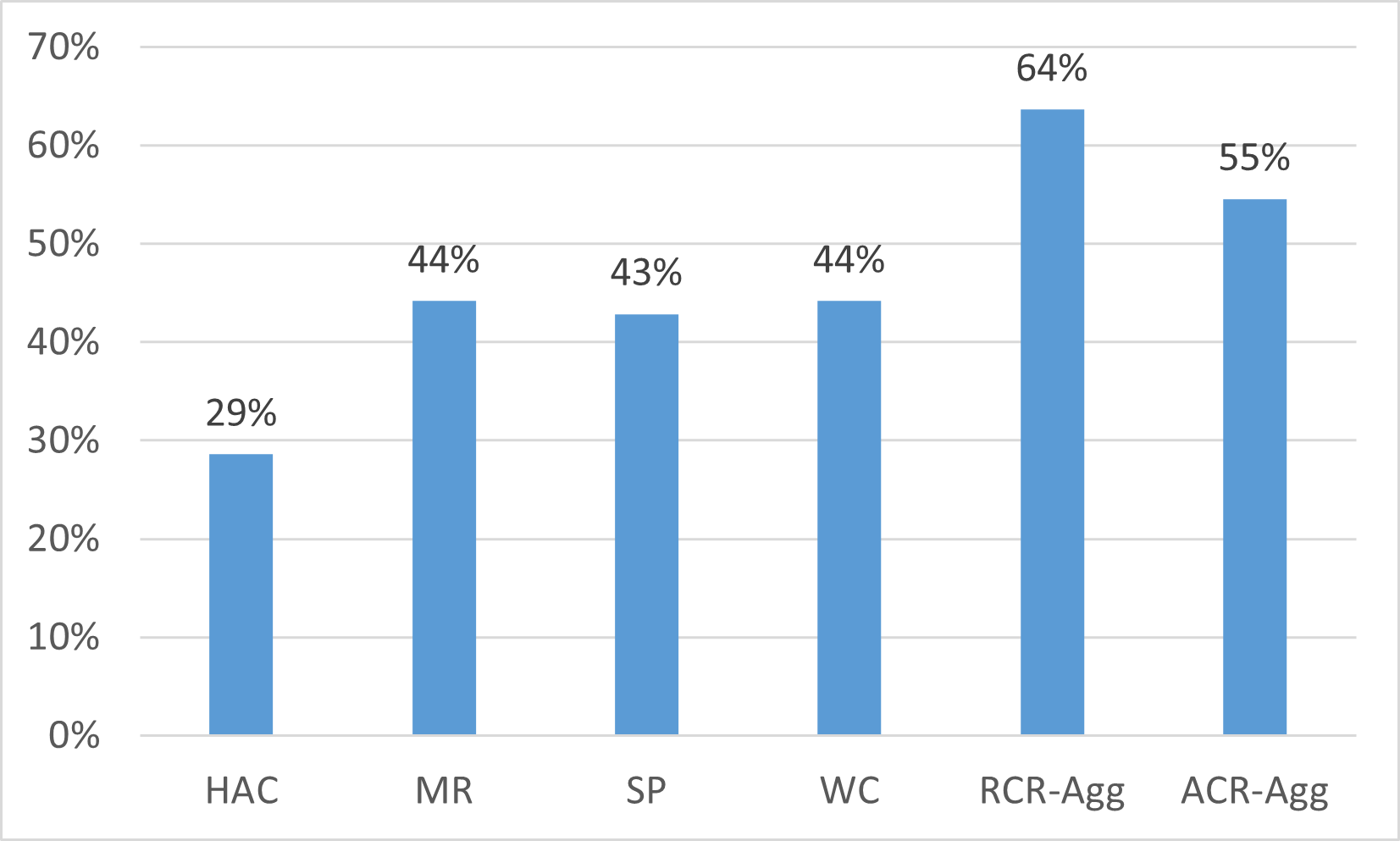}  
	\caption{Aggregation methods' success rate (the percentage of cases in which the chosen answer was the correct one).}
	\label{fig:res}
\end{figure}

Figure \ref{fig:res} presents the success rate, i.e., percentage of cases where the correct answer was identified, of both the standard and new aggregation methods, over the sub-sets of responses (${R_P}'\subset R_P$) for all $P$. 
As can be seen, the success rate of our proposed aggregation methods, RCR-Agg and ACR-Agg, which were 64\% and 55\%, respectively, surpassed all other methods.
The best ML-Based method, RCR-Agg has a significantly higher success rate than the best standard method.

To demonstrate the results' statistical significance we performed a McNemar match-pairs sign test, comparing the success of the ML-based methods (RCR-Agg and ACR-Agg) to each rule-based method, over all instances in the data set. For both RCR-Agg and ACR-Agg, the results indicate a significant increase in successes in every comparison (HAC vs RCR-Agg, $p<4.896E-05$; MR vs RCR-Agg, $p<0.00031$; SP vs RCR-Agg, $p<0.0014$; WC vs RCR-Agg, $p<0.00031$; HAC vs ACR-Agg $p<0.00266$; MR vs ACR-Agg, $p<0.02686$; SP vs ACR-Agg, $p<0.0265$; WC vs ACR-Agg, $p<0.02686$).

\begin{figure}[h]
	\centering
  \includegraphics[width=0.9\columnwidth]{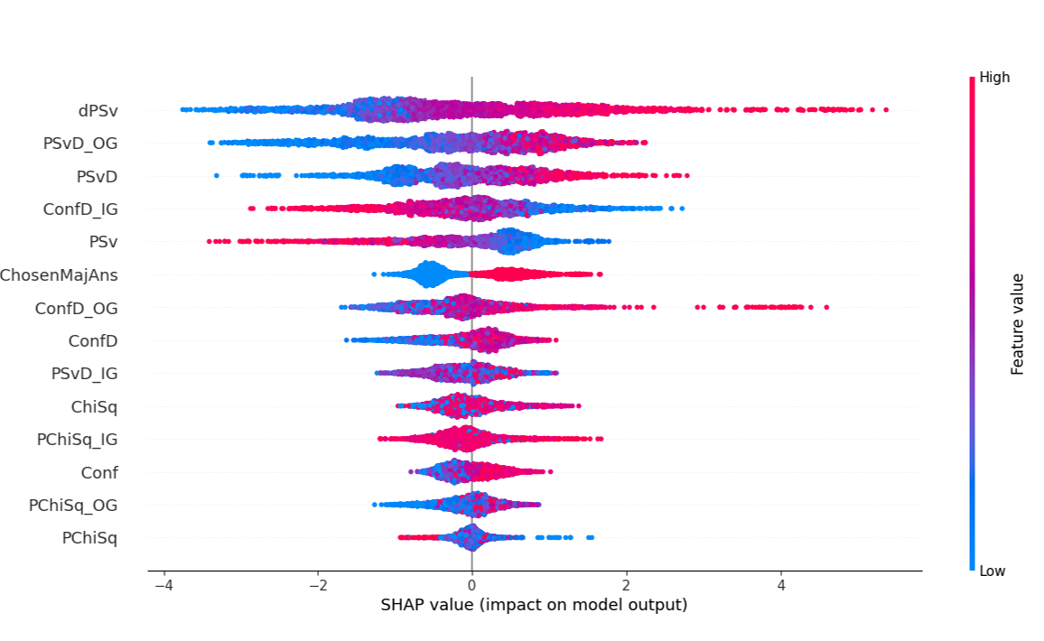}  
  \caption{SHAP features evaluation: Responses-based classification model that uses the RCR approach}
  	\label{fig:shap-sub}
\end{figure}

\begin{figure}[h]
	\centering
  \includegraphics[width=0.9\columnwidth]{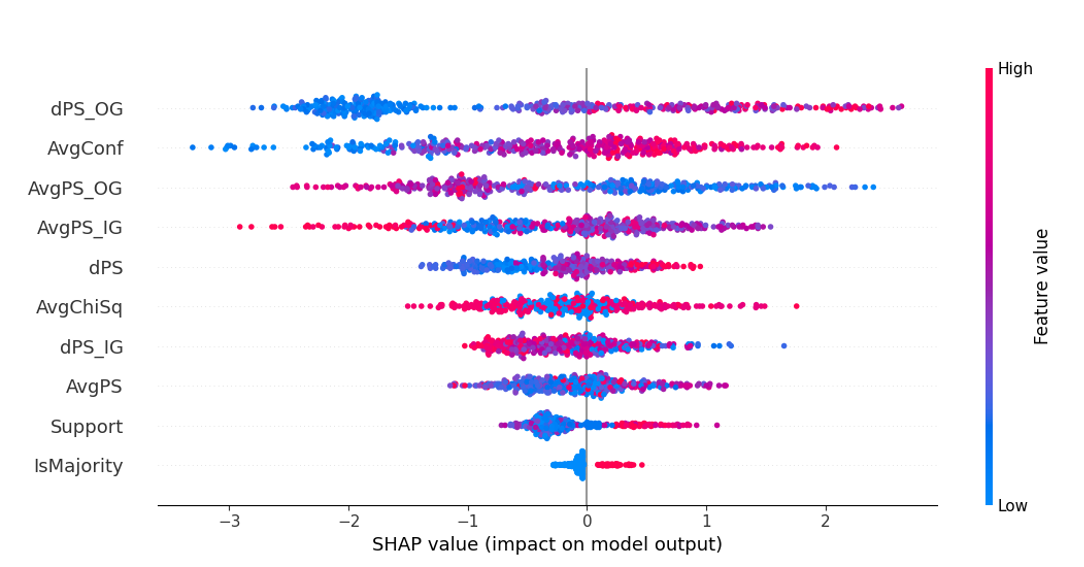}  
  \caption{SHAP features evaluation: Answers-based classification model that uses the ACR approach}
  	\label{fig:shap-ans}
\end{figure}

\subsection{Evaluation of the Features Importance using SHAP values}

We assessed  the relative importance of the features used by the RCR and ACR feature-engineering approaches, namely, their respective contributions to the performance of the responses-based and the answers-based classification models. To do that, we used a popular feature evaluation method that provides several types of insight through the computation of the SHapely Additive exPlanation (SHAP) values \citep{lundberg2017unified}. Figures \ref{fig:shap-sub} and \ref{fig:shap-ans} details the SHAP evaluation results for the responses-based and the answers-based classification models, respectively. 

We first analyze the SHAP values for the classifier of the responses that uses the RCR feature-engineering approach (Figures \ref{fig:shap-sub}). For starters, the most influential feature is \textit{dPSv}, representing the individual level information used by the \textit{Surprisingly Popular} (SP) answer rule-based aggregation method, which compares the predicted support of $v$ (i.e., the respondent's chosen answer) to its actual support. 
Actually, the top three features with the most impact on this model's output, all originated from the predicted support of $v$. The importance ranking emphasizes the key role of this raw meta-cognitive information, when attempting to identify correct responses. 

Another interesting observation from the SHAP evaluation is the differences between the features applied to the responses of the \textit{in-group}  members (i.e., the supporters of the answer) vs. to the responses of the \textit{out-group} members (i.e., non-supporters of the answer) at the responses-based model (Figure \ref{fig:shap-sub}). For example, the feature \textit{PSvD${_{OG}}$}, which provides the distance of predicted support of the respondent's voted answer from the average prediction of the out-group, has a high impact on the models' predictions, such that high values are more associated with correct responses. However, note that the same information, when applied to the in-group, as provided by the feature \textit{PSvD${_{IG}}$}, has a much lower impact. 

Focusing on the confidence-based features, the basic feature, \textit{Conf}, which provides the raw value of reported confidence, has a fairly low impact on the model's classification. This can be expected due to the unimpressive performances of the confidence-based aggregation methods, HAC and WC, as shown in Figure \ref{fig:res}. This is also consistent with understating the importance of context when referring to confidence reports, as it can be an unreliable indication of the respondent's abilities. That being said, observing the individual confidence level relative to other responses, as in \textit{ConfD${_{IG}}$} and \textit{ConfD${_{OG}}$}, can provide the context needed to elevate its value. And indeed, the SHAP evaluation shows that these features have a high impact on the responses-based classifications, albeit in an \textit{inverse} direction (a \textit{higher} confidence relative to the in-group is associated with a \textit{lower} probability of correctness). Note that contrary to the features originating from the predicted support of $v$, here the comparison to the in-group (\textit{ConfD${_{IG}}$}) is more informative than the comparison to the out-group (\textit{ConfD${_{OG}}$}).
This difference is another demonstration of the benefit of extracting information by dividing the respondents into sub-groups based on their answers. By that, we can achieve a deeper knowledge that assists in identifying solvers. 

If we turn our focus to the analysis of the SHAP values for the classifier of the answers that uses the ACR feature-engineering approach 
(Figure \ref{fig:shap-ans}), we see that the most valuable feature (\textit{dPS${_{OG}}$}) consists of the distance between the actual answer's support and the support predicted by its non-supporters (the respondents who did not vote for that answer). This analysis revealed a possible new aggregation method that uses this value as its aggregation rule; it might be viewed as an extension of the Surprisingly Popular Option method for multiple-choice questions.
We see that the SHAP analysis of the methodology based on using the ACR is aligned with the one based on the use of the RCR, as in both cases the most valuable feature is a function of the predicted support of $v$, emphasizing the importance of this information in the aggregation process. 
Another resemblance is in the comparison between observing the in-group-based features vs the out-group-based features, as again we see that it is  more informative to observe the support predicted by the out-group than the support predicted by the in-group.

\section{Conclusions and Future Work}

The results of the current study demonstrate the highly encouraging potential of using our proposed feature engineering process, including meta-cognitive features, to train a machine-learning classification model for identifying correct and incorrect answers by a group of respondents. 

We observed a significant improvement when using the ML-based methods, especially the RCR-Agg method. The fact that the RCR representation methodology outperformed the ACR method could be due to a more refined level of resolution offered when analyzing the multiple individuals and their responses, compared to the group (as in RCR), as opposed to the more aggregated nature of the features of the small number of answers when considering the entire group responses, or the influence of the small sizes of the answers-based data sets (due to the small number of possible answers for each task). 
That being said, there might be a group of problems that have a very broad amount of solutions, to which the ACR approach might potentially be more suited than RCR.  

To summarize, our design and evaluation of the ML-based aggregation methods makes two contributions. 
First, these methods seemingly capitalized on the basic aggregation rule of the baseline methods and replicated it, when needed, i.e., when the baseline methods succeeded, so did the ML-based methods, in the vast majority of the time.
Second, these methods managed to "compose" new aggregation-rules for identifying the correct answer, in the cases in which the rules used by the standard baseline methods did \textit{not} succeed.  

We also tested the option of using an answer-subject combination voting method, which bases its decision on the results of \textit{both} the answer-based classifier and the subject-based classifier. However, although this methods' performance did surpass the performance of the ACR-Agg, it did not surpass  the performance of the RCR-Agg, at least in the context of the  problems currently included in our study. To simplify the presentation of the results, we did not include this algorithm in our analysis. We intend to focus on improving the ACR-Agg method, with respect to its various aspects, as well as continue our attempt to develop a successful integration method which bases its decision upon an \textit{ensemble} of different types of classification approaches (e.g., responses-based and answers-based). 

Finally, a potentially promising direction to investigate, assuming a sufficiently large data set of problems, responses, and answers, is the application of deep-learning techniques for generating valuable features for the different types of classification models. The result might increase the accuracy of the machine-learning models, and thus, can increase the success rate of aggregation methods which base their decisions upon the models' classifications. 

\bibliographystyle{unsrtnat}
\bibliography{references}  






\end{document}